# Scheduling a Dynamic Aircraft Repair Shop with Limited Repair Resources

**Maliheh Aramon Bajestani**                 MARAMON@MIE.UTORONTO.CA
**J. Christopher Beck**                     JCB@MIE.UTORONTO.CA
*Department of Mechanical & Industrial Engineering*
*University of Toronto, Canada*

## Abstract

We address a dynamic repair shop scheduling problem in the context of military aircraft fleet management where the goal is to maintain a full complement of aircraft over the long-term. A number of flights, each with a requirement for a specific number and type of aircraft, are already scheduled over a long horizon. We need to assign aircraft to flights and schedule repair activities while considering the flights requirements, repair capacity, and aircraft failures. The number of aircraft awaiting repair dynamically changes over time due to failures and it is therefore necessary to rebuild the repair schedule online. To solve the problem, we view the dynamic repair shop as successive static repair scheduling sub-problems over shorter time periods. We propose a complete approach based on the logic-based Benders decomposition to solve the static sub-problems, and design different rescheduling policies to schedule the dynamic repair shop. Computational experiments demonstrate that the Benders model is able to find and prove optimal solutions on average four times faster than a mixed integer programming model. The rescheduling approach having both aspects of scheduling over a longer horizon and quickly adjusting the schedule increases aircraft available in the long term by 10% compared to the approaches having either one of the aspects alone.

## 1. Introduction

The United States Air Force website outlines readiness as one of the vital elements to ensure the operational effectiveness in defense strategies (Schwartz, 2012). Deficiency in resourcing for operations and maintenance is stated as the common culprit for poor readiness. Since air forces are budget-constrained and highly dynamic environments, the optimal allocation of resources to activities to maintain readiness at an appropriate level is challenging. In this paper, we study a problem in the context of military aircraft repair shop where readiness is defined as the ability to effectively carry out the pre-scheduled missions. We need to decide when each failed aircraft should be repaired to guarantee the availability of aircraft at high and steady level. However, high frequency of unexpected failures in military aircraft (Safaei, Banjevic, & Jardine, 2011) and limited repair resources such as workforce, tools, and space (Kozanidis, Gavranis, & Kostarelou, 2012) constrain consistent aircraft availability.

Motivated by the work of Safaei et al. (2010, 2011), we study the problem of scheduling a military aircraft repair shop, where a number of flights are planned over a long horizon. Every flight, also called *a wave*, has a maximum requirement for a specific number of aircraft of different types though it can be partially carried out without its maximum complement. At the beginning of the time horizon, an aircraft is either ready for a pre-flight check or is awaiting repair in the repair shop. Aircraft flow over a long horizon is illustrated in Figure 1. The goal is to determine an assignment of aircraft to waves and a schedule of repair jobs that will maximize flight coverage, that is, the extent to which the aircraft requirements of the flights are met. When an aircraft fails during a pre- or post-flight check, it enters the repair shop to be incorporated into the current repair schedule. Each aircraft failure requires a set of independent repair activities with known characteristics such as processing times and resource requirements to be scheduled on repair resources with a limited capacity.

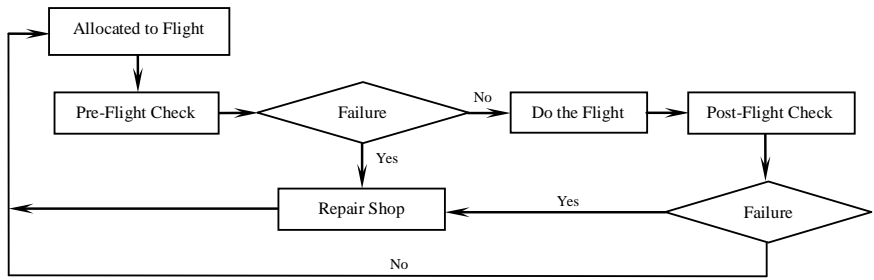

Figure 1: Aircraft flow among waves, checks, and the repair shop over a long horizon.

The central idea of our solution approach is to view the dynamic repair shop as successive static sub-problems over shorter time periods. A solution of the static sub-problem determines an assignment of aircraft to flights and a schedule of repair jobs maximizing the flight coverage. When a failed aircraft enters the repair shop while the previous repair schedule is still under execution, we reschedule the repair activities by solving a new static sub-problem.

After proving the NP-hardness of the static sub-problem, we explore several techniques to solve the problem: mixed integer programming (MIP); constraint programming (CP); logic-based Benders decomposition (LBBD) using either MIP or CP; and a dispatching heuristic motivated by the Apparent Tardiness Cost (ATC) dispatching rule. We then design three different rescheduling policies based on the length of the scheduling horizon and how frequently rescheduling is done.

We perform two separate empirical studies. The first indicates that the integration of the dispatching heuristic and LBBD results in the lowest mean run-time of the techniques tested to optimally schedule the repair shop. The second experiment demonstrates that both defining the static scheduling problem over a longer horizon and rescheduling more frequently provide the flights with 10% higher coverage than either one of them alone.

The remainder of the paper is organized as follows: We define the problem, and provide an overview of the relevant literature in Section 2. Section 3 proves the NP-hardness of the static sub-problem, defines a number of solution approaches for it, presents the details of the proposed policies for rescheduling the dynamic repair shop, and describes our model of the aircraft failures. The computational results on the performance of different scheduling techniques and on how and when rescheduling should be done are described in Section 4. A discussion of our solution approach and results are presented in Section 5. We end with conclusion and directions for the future work in Section 6.

## 2. Background

In this section, the formal definition of the problem is given and the relevant literature is reviewed.

### 2.1 Problem Definition

Figure 2 is a snapshot of the problem at time 0, where circles represent aircraft. A number of flights (five are shown) and their corresponding pre- and post-flight checks are already scheduled over a long horizon. It is assumed that the total number of aircraft is constant over a long horizon. A number of aircraft (three in the diagram) are ready for the pre-flight check while others are currently in the shop awaiting repair before they can proceed to a pre-flight check. Failure is only detected during a check and we assume that a check will always correctly assess the status of an aircraft at negligible cost and that the duration of a check is incorporated in the length of the corresponding wave.

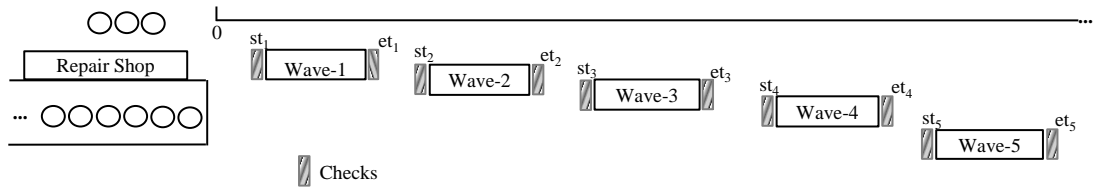

Figure 2: Snapshot of the problem at time 0 over a long horizon.

The goal is to assign aircraft to waves to maximize coverage while at the same time creating a feasible repair schedule. The scheduling problem is under the constraints that the repair shop has limited capacity and the aircraft are subject to breakdown. We assume that once an aircraft fails, it goes to the repair shop and waits until its repair operations are performed.

We use the following notation to represent the problem.

- $N$ is the set of aircraft. $\lambda_n$ is the failure rate of the aircraft $n \in N$ denoting the frequency of failure per time unit. For example if the failure rate is 0.2 per day, it means that the mean time to aircraft failure is 5 days.

- $K$ is the set of aircraft types. $I_k$ denotes the set of aircraft type $k \in K$ where $\eta_k$ aircraft are ready (i.e., not in the repair shop at time 0). Let $|I_k|$ denote the number of aircraft of type $k$, $|I_k| - \eta_k$ aircraft of type $k$ are then in the repair shop at time 0. $\bar{\lambda}_k$ is the mean failure rate over all aircraft of type $k$.

- $R$ is the set of repair resources (called *trades*). The maximum capacity of trade $r \in R$ is $C_r$.

- $W$ is the set of waves. Each wave, $w \in W$, has a start-time, $st_w$, and an end-time, $et_w$. Each wave requires at most $a_{kw}$ aircraft of type $k$.

- $J$ is the set of existing jobs in the repair shop. Each job is associated with a specific aircraft type. $M_r$ is the set of jobs requiring trade $r$. Each job might require more than one trade to be completed. The processing time of job $j$ on trade $r$ is $p_{jr}$ and $c_{jr}$ is the capacity of trade $r$ required by job $j$.

To model the deterioration of an aircraft, each time it flies a wave its failure rate, $\lambda_n$, increases by $\gamma$ percent, i.e., its failure rate is $(1 + 0.01 \times \gamma)\lambda_n$ after the flight. If an aircraft fails, its failure rate after repair returns to what it was just before the failure. In the other words, as in one of the standard repair models in the maintenance literature, repair is minimal (Wang, 2002). The probability of diagnosing aircraft $n$ as failed in pre- and post-flight checks is a function of its failure rate right before the checks denoted as $f^{pre}(\lambda_n)$ and $f^{post}(\lambda_n)$. The probability of failure detection in pre-flight checks is smaller than the post-flight checks because an aircraft is either just released from the repair shop or has already passed a previous post-flight check successfully (Safaei et al., 2011).

To find the probability of failure of an aircraft in pre- and post-flight checks for a specific wave, we need to track the complete history of the aircraft. For example, if we assume that a given aircraft is repaired and assigned to the first wave, then there are three paths: the aircraft fails the pre-flight check; the aircraft passes the pre-flight check, flies the wave, and fails the post-flight check; or the aircraft passes the pre-flight check, flies the wave, and passes the post-flight check. Therefore, the availability of the aircraft for the second wave can be represented as a random variable whose expected value depends on the probability of these three different paths and the scheduling decisions to repair the failed aircraft before the second wave. Similarly the availability of the aircraft for subsequent waves depends on its entire path through the checks, repair shop, and waves. As the number of waves and aircraft increase, the size of the state space will become prohibitive. Furthermore, the repair scheduling decisions themselves impact the aircraft histories: the probability that an aircraft is available for the third wave is different depending on if it was repaired in time for the first wave or only for the second wave. The details on approximating the failure probabilities are presented in Section 3.2.1.

As the complexity of the problem has not been shown, we prove that it is NP-hard in Section 3.1.

## 2.2 Literature Review

This section provides necessary background on repair shop scheduling problem and the logic-based Benders decomposition.

### 2.2.1 Repair Shop Scheduling

Repair shops have been mainly studied as a machine-repairman problem (Haque & Armstrong, 2007; Stecke, 1992) which has a set of workers and a set of machines that are subject to failures and therefore need repair. Workers and machines respectively correspond to trades and aircraft, in our problem. As the number of workers is less than the number of machines, it is necessary to allocate the repair jobs to the workers with the goal of optimizing a given performance measure (e.g., the total expected machine downtime) over the long-term. Derman, Liberman, and Ross (1980) did the early work on solving the scheduling problem of a repair shop with a single repairman. They showed that repairing the failed machines in non-decreasing order of failure rate stochastically maximizes the number of working machines. The literature on the scheduling of a repair shop was then extended by considering multiple repairmen, preemptive and non-preemptive repair, and different failure and repair distributions. A comprehensive review of the literature on the scheduling of a repair system is provided by Iravani, Krishnamurthy, and Chao (2007).

The analytical models in the literature are mainly developed using Markov Decision Processes (dynamic programming) and guarantee the optimality of a given performance measure in the long-term. These models often do not consider the combinatorics of the real scheduling problems such as different repair capacity limits, different due dates, and different resource and processing requirements. Therefore, they typically result in a static dispatching-type repair policy similar to that found by Derman et al. (1980). However, in our problem, the waves have different plane requirements and the processing times and the resource requirements of the repair activities become known when they enter the repair shop. Therefore, we believe that a better performance can be achieved by dealing directly with the combinatorics and explicitly scheduling the repair shop to meet the waves. To handle the uncertain and combinatorial structure of the scheduling problems, we use a dynamic scheduling approach.

Dynamic scheduling is the methodology developed in the scheduling literature where the operational uncertainties like machine breakdowns or the unexpected arrival of new orders prevent the execution of the schedule as planned (Aytug, Lawley, McKay, Mohan, & Uzsoy, 2005; O'Donovan, Uzsoy, & McKay, 1999). The dynamic scheduling problem is often viewed as a collection of linked static sub-problems. Taking this view makes the myriad of algorithms developed for the static scheduling problems applicable. The developed algorithms can deal with the combinatorics of the scheduling problems and can optimize the quality of the schedule at discrete time points, i.e., for each static sub-problem. However, as they cannot completely deal with the operational uncertainties, a real-time disruption requires modification of the schedule to either permit the execution or to improve the quality of the schedule considering the most recently revealed information. The process of modifying the previous schedule is called rescheduling (Vieira, Hermann, & Lin, 2003; Aytug et al., 2005; Bidot, Vidal, Laborie, & Beck, 2009). How to solve static sub-problems and how to connect them using rescheduling strategies are the main aspects of the dynamic scheduling research. Although we are not aware of using dynamic scheduling in a repair system, it has been successfully applied to a variety of scheduling problems including single machine (Ovacik & Uzsoy, 1994; O'Donovan et al., 1999; Cowling & Johansson, 2002), parallel machines (Vieira, Hermann, & Lin, 2000; Ovacik & Uzsoy, 1995), and job shop (Sabuncuoglu & Bayiz, 2000; Liu, Ong, & Ng, 2005; Vinod & Sridharan, 2011).

Other areas of literature with similarities to our static problem are the operational level maintenance scheduling problem, in general, and the flight and maintenance planning problem of military aircraft, specifically. The former literature addresses the problem of finding a schedule for given maintenance activities such that the sum of maintenance costs is minimized. The focus is on the operational level, determining the maintenance activities performed in each time period (Budai, Huisman, & Dekker, 2006). Starting with the early work of Wagner, Giglio, and Glaser (1964), this literature was extended through developing mathematical models and effective solution approaches for a variety of applications (Frost & Dechter, 1998; Haghani & Shafahi, 2002; Budai et al., 2006; Grigoriev, van de Klundert, & Spieksma, 2006). The latter literature studies the problem of maintenance planning and mission assignment of military aircraft where the goal is to decide which aircraft to fly and which one to perform maintenance on, maximizing their long-term availability. Similar to the maintenance scheduling literature, mathematical programming is the common approach to solve the problems of this literature. Kozanidis, Gavranis, and Kostarelou (2012) recently proposed a mixed integer non-linear programming model to optimize the joint flight and maintenance plan of mission aircraft.

Safaei et al. (2010) modeled the static problem addressed here as an operational level maintenance scheduling problem using MIP. Their MIP includes an assignment problem and two network problems: the former assigns the aircraft to the waves and the latter calculates the expected number of available aircraft for the waves as well as the expected number of available workers for the repair jobs. They later extended their work by using slightly different MIP model where the time-indexed approach is used to enforce the workforce availability constraint and they verified the validity of their model by a number of instances under different combinations of workforce sizes (Safaei et al., 2011).

The difference between the static problem addressed in this paper and the previous works on operational maintenance scheduling is that our objective function (flight coverage) depends not only on the scheduling decisions but also on the outcomes of the pre- and post-flight checks. These two quite different components of the problem motivate the decomposition approach, logic-based Benders decomposition, reviewed below.

### 2.2.2 Logic-Based Benders Decomposition

The classical Benders decomposition (Benders, 1962; Geoffrion & Graves, 1974) is a mathematical programming approach for solving large-scale mixed integer programming models. It partitions the problem into a mixed integer master problem (MP) which is a relaxation of the global model and a set of linear sub-problems (SPs). Solving a problem by classical Benders involves iteratively solving the MP to optimality and using the solution to generate the sub-problems. The linear programming dual of the SPs is then solved to derive the tightest bound on the global cost function. If this bound is greater than or equal to the current MP solution (assuming a maximization problem), the MP solution and the SP solutions constitute a globally optimal solution. Otherwise, a constraint, a *Benders cut*, is added to the MP to express the violated bound and another iteration is performed.

Logic-based Benders decomposition (Hooker & Yan, 1995; Hooker & Ottosson, 2003) was developed excluding the necessity that the MP should be a mixed integer model and the SPs should be linear. Therefore, the inference duals (Hooker, 2005) of the SPs are solved rather than the linear duals to find the tightest bound on the global cost function from the original constraints and the current MP solution. Although the logic-based Benders decomposition has more flexibility in modeling the problems, there is no standard procedure to derive the Benders cuts.

Representing the relaxation of SPs in the MP, and designing a strong Benders cut are of great importance in decreasing the computational effort to identify the globally feasible and optimal solution. The former results in MP solutions which are likely to satisfy the SPs, and the latter rules out a large number of MP solutions in each iteration (Hooker, 2007).

Logic-based Benders decomposition has been shown to be effective in a wide range of problems including scheduling (Hooker, 2005, 2007; Beck, 2010), facility and vehicle allocation (Fazel-Zarandi & Beck, 2012), and queue design and control problems (Terekhov, Beck, & Brown, 2009).

## 3. Solution Approach

The main idea of our solution approach is to view the dynamic problem as linked successive static sub-problems. As noted above, this is a common approach in dynamic scheduling. This view results in a rescheduling strategy based on scheduling static sub-problems over shorter time periods. Therefore, we have two sub-goals: how to solve and how to connect the static sub-problems. In this section, we first show that the static problem is NP-hard and present different solution techniques for solving it. We then define three rescheduling strategies designed to connect the static sub-problems. Finally, we describe our approach for modeling the dynamic events, i.e., aircraft failures.

### 3.1 The Complexity of the Static Repair Shop Problem

We establish the NP-hardness of the static repair shop problem by reduction from a single machine scheduling problem with the objective of minimizing the weighted number of tardy jobs and with a common due date, i.e., $1|d_j = d|\sum w_j U_j$[1] where $d_j$, $w_j$, and $U_j$ denote the due date, the weight, and the variable representing whether a job is tardy or not for job $j$, respectively. Note that job $j$ is tardy if its processing is not finished at or before its due date. This problem is equivalent to a one dimensional knapsack problem with non-uniform profit which is shown to be NP-complete by reduction from the PARTITION problem (Pinedo, 2002).

**Theorem 1.** *The static problem is NP-hard.*

*Proof.* We show that an instance of the single machine scheduling problem, $I$, with a common due date and with the objective of minimizing the weighted number of tardy jobs reduces to the static repair shop problem. In instance $I$, we assume that there are $T$ jobs, the $j$th having a processing time $t_j$, a weight $w_j$, and a common due date $d$. Without loss of generality, we further assume that the weights, $w_j$, are in the interval $[0, 1]$. The objective is to schedule the jobs such that the sum of the weights of the tardy jobs is minimized or equivalently the sum of the non-tardy job weights is maximized. From instance $I$, an instance of the static repair shop problem can be formulated such that there is one wave, there are $T$ failed aircraft in the repair shop ($|N| = T$), there are $T$ aircraft types ($|K| = T$), and there is one repair resource ($|R| = 1$) with capacity $C = 1$. The start-time of the wave is $st_1 = d$, requiring all $T$ aircraft. Each failed aircraft, $j$, has a different type, and corresponds to one repair job in the repair shop with the processing time $p_{j1} = t_j$ and resource requirement $c_{j1} = 1$ for the single resource. The probability of failure in the pre-check of the wave for aircraft $j$ is $(1 - w_j)$. The repaired aircraft $j$ contributes to the flight coverage if it survives the pre-check with probability $w_j$. Therefore, to maximize the flight coverage, the goal is to schedule the failed aircraft so that the sum of the probabilities that the repaired aircraft survives the pre-check, i.e., $\sum w_j$ is maximized. The goal is equivalent to maximizing the sum of the non-tardy job weights in instance $I$. As the single machine scheduling problem with the objective of the weighted number of tardy jobs and with a common due date, i.e., $1|d_j = d|\sum w_j U_j$, is NP-hard (Pinedo, 2002), we conclude that the static repair shop problem is also NP-hard.                     □

---

1. The notation used for describing a problem in scheduling literature is $\alpha|\beta|\gamma$ where $\alpha$ represents the machine environment, $\beta$ describes the processing characteristics and constraints in detail, and $\gamma$ denotes the objective function (Pinedo, 2002).

## 3.2 Scheduling Techniques

We investigate a number of approaches to solve the repair shop scheduling problem including mixed integer programming, constraint programming, logic-based Benders decomposition, a dispatch rule, and a simple hybrid approach. Each of the approaches is described in detail in this section.

### 3.2.1 Mixed Integer Programming

Mixed integer programming (MIP) is the default solution approach for many scheduling problems (Heinz & Beck, 2012). In a MIP formulation, the constraints are represented in the form of linear equalities and/or inequalities and polyhedral theory and linear programming techniques such as relaxation and cutting planes embedded in the state-of-the-art MIP solvers are applied to solve the problem (Queyranne & Schulz, 1994; Heinz & Beck, 2012). We propose a novel mixed integer programming model where the uncertainty in the outcome of the checks is modeled as expectation. This model is different from and, as we show below in Section 4.1.2, significantly faster than those of Safaei et al. (2010, 2011). Table 1 summarizes the notation defined in Section 2.1 and defines the decision variables of the MIP model.

In this section, without loss of generality, we interpret $W$ as the set of waves in the current static sub-problem and consider the start-times of the waves as due dates to finish the repair of the aircraft. Therefore, we define $D = \{d_i | i = 1, 2, ..., |W|, |W| + 1\}$ to be an ordered set of due dates consisting of the wave start-times plus a big value, $B$, sorted in ascending order. More specifically, $d_i$ equals to the start-time of the $i$-th wave, $st_i$. Because of the limited repair capacity, it is possible that some of the failed aircraft cannot be repaired in time for any of the waves. In such a case, the due date of the repair job is assigned to $d_{|W|+1} = B$. In our model, $B$ equals the sum of the start-time of the last wave and the maximum processing times of all the jobs over all the trades, i.e., $d_{|W|} + \max_{j,r}(p_{jr})$ and we do not enforce the repair resource capacity after $d_{|W|}$.

As explained in Section 2.1, the exact calculation of the aircraft failure probability and consequently of the expected number of available aircraft is intractable since it depends on the complete aircraft histories. Therefore, we distinguish aircraft based on their type and use a recursive equation (Equation 4) to approximate the expected number of available aircraft. The details of Equation (4) are provided later in this section. For each aircraft type of $k$, the average failure rate, $\bar{\lambda}_k$, is used to calculate the probability of failure during pre- and post-flight checks, respectively: $\xi_k^{pre} = f^{pre}(\bar{\lambda}_k)$ and $\xi_k^{post} = f^{post}(\bar{\lambda}_k)$. Furthermore, the failure rate of each aircraft is assumed to remain constant in the scheduling horizon of the static problem and to not increase after flying a wave. Therefore, our approximation is likely to underestimate the number of actual aircraft failures.

| Notation | |
|---|---|
| $N = \{1, ..., n, ..., |N|\}$ | The set of aircraft |
| $K = \{1, ..., k, ..., |K|\}$ | The set of aircraft types |
| $R = \{1, ..., r, ..., |R|\}$ | The set of trades |
| $W = \{1, ..., w, ..., |W|\}$ | The set of waves |
| $J = \{1, ..., j, ..., |J|\}$ | The set of repair jobs (failed aircraft) in the repair shop |
| $\lambda_n$ | The failure rate of aircraft $n$ |
| $I_k$ | The set of the aircraft of type $k$ |
| $\eta_k$ | The number of aircraft of type $k$ at the repair shop at time 0 |
| $\tilde{\lambda}_k$ | The average failure rate over all aircraft of type $k$ being equal to $\frac{\sum_{n \in I_k} \lambda_n}{|I_k|}$ |
| $\xi_k^{pre}$ | The probability that aircraft type $k$ fails in pre-flight check |
| $\xi_k^{post}$ | The probability that aircraft type $k$ fails in post-flight check |
| $st_w$ | The start-time of wave $w$ |
| $et_w$ | The end-time of wave $w$ |
| $a_{kw}$ | The maximum number of aircraft of type $k$ required by wave $w$ |
| $M_r$ | The set of the repair jobs requiring trade $r$ |
| $C_r$ | The maximum capacity of trade $r$ |
| $p_{jr}$ | The processing time of job $j$ on trade $r$ |
| $c_{jr}$ | The capacity of trade $r$ required to process job $j$ |
| $D = \{d_1, ..., d_i, ..., d_{|W|+1}\}$ | The set of due dates where $d_i = st_i, \forall i \leq |W|$ and $d_{|W|+1} = B$ |
| $B$ | The big value equal to $d_{|W|} + \max_{j,r}(p_{jr})$ |
| **Decision Variables** | |
| $Z_{kw}$ | The number of aircraft of type $k$ assigned to fly in wave $w$ |
| $x_{ij}$ | $x_{ij} = 1$ if the $i$th due date is assigned to job $j$, and $x_{ij} = 0$ otherwise |
| $st_{jr}$ | The start-time of job $j$ on trade $r$ |
| **Inferred Variables** | |
| $U_{kw}$ | The number of aircraft of type $k$ whose repair due date is $st_w$ |
| $E_{kw}$ | The expected number of available aircraft of type $k$ for wave $w$ |
| $et_{jr}$ | The end-time of job $j$ on trade $r$ |

Table 1: Summary of notation; the decision variables and inferred variables for the MIP model.

The MIP model is shown in Figure 3 where $Z_{kw}$, the number of aircraft of type $k$ that are assigned to fly in wave $w$, is a true decision variable: we can choose to send fewer aircraft on a wave than are currently (in expectation) available. In contrast, $E_{kw}$ is the expected number of aircraft of type $k$ available for wave $w$ and is based on the probabilistic outcomes of previous waves and the number of newly repaired aircraft ($U_{kw}$). We refer to this model as *MIP* and we rely on the default branch-and-bound search in the IBM ILOG CPLEX 12.3 solver, a state-of-the-art commercial MIP solver to solve it. The details of MIP model are summarized as follows:

$$\text{Maximize} \sum_{w=1}^{|W|} \sum_{k=1}^{|K|} Z_{kw} \tag{1}$$

Subject to:

$$U_{kw} = \sum_{j \in I_k, \ i=w} x_{ij}, \qquad\qquad \forall k, \ \forall w \tag{2}$$

$$E_{k1} = (\eta_k + U_{k1})(1 - \xi_k^{pre}), \qquad\qquad \forall k \tag{3}$$

$$E_{kw} = (E_{k(w-1)} - Z_{k(w-1)} + U_{kw})(1 - \xi_k^{pre})$$
$$+ \sum_{v \in \mathcal{V}_w} Z_{kv}(1 - \xi_k^{post})(1 - \xi_k^{pre}), \qquad\qquad \forall w(w \neq 1), \ \forall k \tag{4}$$

$$Z_{kw} \leq a_{kw}, \qquad\qquad \forall k, \ \forall w \tag{5}$$

$$Z_{kw} \leq E_{kw}, \qquad\qquad \forall k, \ \forall w \tag{6}$$

$$\sum_{i=1}^{|W|+1} x_{ij} = 1, \qquad\qquad \forall j \tag{7}$$

$$st_{jr} + p_{jr} = et_{jr}, \qquad\qquad \forall j, \ \forall r \tag{8}$$

$$et_{jr} \leq \sum_{i=1}^{|W|+1} x_{ij} d_i, \qquad\qquad \forall j, \ \forall r \tag{9}$$

$$\sum_{j \in M_r} c_{jr}((t \geq st_{jr}) \land (t < et_{jr})) \leq C_r, \qquad\qquad \forall t(t \leq st_{|W|}), \ \forall r \tag{10}$$

$$x_{ij} \in \{0, 1\}, \qquad\qquad \forall i, \ \forall j \tag{11}$$

$$0 \leq E_{kw} \leq |N|, \qquad\qquad \forall k, \ \forall w \tag{12}$$

$$st_{jr}, et_{jr} \in \mathbb{Z}^+ \cup \{0\}, \qquad\qquad \forall j, \ \forall r \tag{13}$$

$$Z_{kw} \in \mathbb{Z}^+ \cup \{0\}, Z_{kw} \leq |N|, \qquad\qquad \forall k, \ \forall w \tag{14}$$

Figure 3: The global MIP model for the static repair shop scheduling problem.

- The objective function (1) maximizes the number of aircraft assigned to waves. Although we have modeled the uncertain outcome of the flight checks as expectation, the objective function is not the expected wave coverage because (i) each wave has specific upper bounds on plane requirements and (ii) the maximum wave coverage for each wave is 1. If the expected number of available aircraft, $E_{kw}$, is more than the requirement, $a_{kw}$, for a given wave, the extra aircraft do not fly the wave and so do not contribute to the coverage. By not flying "extra" planes, we do not decrease the probability that they will be available for the next wave.

- Equation (2) calculates the number of aircraft of type $k$ whose repair due date is $st_w$. In the other words, summing the decision variables $x_{ij}$ where job $j$ is an aircraft of type $k$ and where the $i$-th due date corresponds to the start-time of wave $w$ gives the number of aircraft type $k$ leaving the repair shop right before the pre-flight check of wave $w$.

- Equation (3) calculates the expected number of available aircraft of type $k$ for the first wave.

- Equation (4) calculates the expected number of available aircraft of type $k$ for the other waves. The first term includes those aircraft available but not used for the previous wave, i.e., $(E_{k(w-1)} - Z_{k(w-1)})$, and those newly arrived from the repair shop, i.e., $U_{kw}$. The second term sums over all aircraft that become available because they have completed waves since the previous wave started where $\mathcal{V}_w = \{v | v \in W, st_{w-1} < et_v \leq st_w\}$.

- Constraints (5) and (6) ensure that the number of aircraft that are assigned to fly in each wave is less than or equal to the number of aircraft required and the expected number available.

- Constraint (7) assigns exactly one due date to each job.

- Equation (8) calculates the end-time of the jobs.

- Constraint (9) guarantees that the end-time of each job is less than or equal to its assigned due date.

- Constraint (10) is a logical-and constraint enforcing the capacity limit of trade $r$ by summing over the capacity required by the set of jobs under repair at time $t$. Since the jobs after the start-time of the last wave do not contribute to the coverage, the capacity constraint is enforced only until the start-time of the last wave, i.e., $st_{|W|}$. The logical "$\wedge$" constraint evaluates to 1 if and only if its two component constraints both evaluate to 1 and to 0 otherwise. A logical inequality evaluates to 1 if and only if it is true and 0 otherwise. For example, if job $j$ is under repair at time $t$, both logical inequalities, $(st_{jr} \leq t)$ and $(t < et_{jr})$ evaluate to 1 and the logical-and constraint, therefore, evaluates to 1. To linearize this constraint, we rely on the default approaches in IBM ILOG CPLEX for handling logical constraints. These approaches translate the logical constraints into their equivalent linear counterparts by creation of new variables and constraints (CPLEX, 2011).

- Constraints (11) to (14) define the domains of the decision variables.

### 3.2.2 Constraint Programming

Constraint programming is a paradigm for solving combinatorial optimization problems. The success of constraint programming (CP) in solving a wide variety of scheduling problems is well established in the literature (Beck, Davenport, Davis, & Fox, 1998; Baptiste, Pape, & Nuijten, 2001). The scheduling problems are usually defined as one or several instances of the constraints satisfaction problem (CSP) (Baptiste et al., 2001). An instance of CSP can be formally described as a triple of $(V, D, C)$ where $V = \{V_1, V_2, ..., V_n\}$ is a set of $n$ variables, $D = \{D_1, D_2, ..., D_n\}$ is a set of the variable domains, $D_i$ corresponding to the possible values that $V_i$ can take, and $C = \{C_1, C_2, ..., C_m\}$ is a set of $m$ constraints, each defined over a subset of variables. A constraint $C_k = \{V_i, ..., V_j\}$ is defined on the Cartesian product of the domains of the variables in its scope $D_i \times ... \times D_j$ and is satisfied if the assignment of the variables in its scope corresponds to one of the value tuples in the constraint relation (Beck, 1999). Representing the scheduling problems using CSPs results in more modeling flexibility compared to mixed integer programming models as there is no restriction on the type of decision variables and constraints.

CP solves scheduling problems by applying three general tools of heuristic search, constraint propagation, and backtracking within a branch-and-bound search tree (Beck, 1999; Beck & Refalo, 2003). Constraint propagation, one of the key principles contributing to the success of CP, is exploited by representing the problem as a conjunction of global constraints, each of which embeds efficient inference techniques that reduce the solution space within a branch-and-bound search tree (Baptiste et al., 2001). The possibility of using two global constraints, *cumulative* and *global cardinality*, to formulate the static repair shop problem motivates the CP model shown in Figure 4.

To formulate the problem using CP, we use the same decision variables as in Table 1. However, instead of $x_{ij}$, we define $\mathcal{D}_j$ corresponding to the assigned due date for job $j$. The CP model differs from MIP in several constraints defined below.

The global cardinality constraint (gcc) has the syntax of gcc($card, value, base$) where $card$, $value$, and $base$ are arrays of variables, values, and variables, respectively. The gcc constraint is satisfied if $value[i]$ is taken by $card[i]$ elements of $base$. In our CP model, for each aircraft type $k$, Constraint (15) enforces that $U_{kw}$ counts the number of times that the start-time of wave $w$ is assigned as a due date to the jobs associated with a failed aircraft of type $k$.

The cumulative constraint has the syntax of cumulative($s, p, c, C$) where $s = \{s_1, s_2, ..., s_n\}$, $p = \{p_1, p_2, ..., p_n\}$, and $c = \{c_1, c_2, ..., c_n\}$ are vectors of the start-time variables, the processing time values, and the amount of required resource of each job, respectively, and where $C$ is the total resource capacity value. The cumulative constraint ensures that the total amount of resource capacity used at any time never exceeds $C$ (Hooker, 2005).

Constraint (17) enforces the time windows: job $j$ on trade $r$ cannot be started later than ($\mathcal{D}_j - p_{jr}$). Constraint (18) defines the domain of the decision variables $\mathcal{D}_j$.

Maximize Objective (1)

Subject to:

Constraints (3) to (6), (12), (14)

$$\text{gcc}([U_{k1}, U_{k2}, ..., U_{k|W|}], [st_1, st_2, ..., st_{|W|}], [\mathcal{D}_{j \in I_k}]), \qquad \forall k \qquad (15)$$

$$\text{cumulative}([st_{jr} | j \in M_r], [p_{jr} | j \in M_r], [c_{jr} | j \in M_r], C_r), \qquad \forall r \qquad (16)$$

$$0 \leq st_{jr} \leq \mathcal{D}_j - p_{jr}, \qquad \forall j, \ \forall r \qquad (17)$$

$$\mathcal{D}_j \in \{st_1, st_2, ..., st_{|W|}, B\}, \qquad \forall j \qquad (18)$$

Figure 4: The CP model for the static repair shop scheduling problem.

We implement this model using IBM ILOG CP Optimizer 12.3 where the default search is used. The start-time variables, $st_{jr}$, are defined by IloIntervalVar objects. To implement the global constraints, we use IloDistribute class for the gcc constraint and IloPulse and IloAlwaysIn functions for the cumulative constraint. Note that, the cumulative constraint is implemented for any time point $t$ until $st_{|W|}$.

### 3.2.3 Logic-Based Benders Decomposition

As the static problem requires making two different decisions, assigning aircraft to the waves and scheduling repair jobs for failed aircraft, a decomposition approach may be well suited. A logic-based Benders decomposition (LBBD) method can be formulated where the master problem assigns aircraft to waves to maximize wave coverage and the sub-problems create the repair schedules given the due dates derived from the master problem solution. We propose four variations: *Benders-MIP* and *Benders-MIP-T*, where the master problems are solved using MIP, the latter with a tighter sub-problem relaxation ("T" stands for tighter); and *Benders-CP* and *Benders-CP-T* with a constraint programming-based master problem. All models use CP for the scheduling sub-problems.

**The Due-Date Assignment Master Problem (DAMP): MIP Model** To formulate the master problem as a MIP model, we use a binary variable $x_{ij}$ for job $j$ and the $i$-th due date with the same meaning as in the global MIP model. A MIP formulation of DAMP is as follows:

$$\text{Maximize Objective (1)}$$
$$\text{Subject to:} \tag{19}$$
$$\text{Constraints (2) to (7), (11), (12), (14)}$$

$$\sum_{j \in M_r} c_{jr} p_{jr} \leq C_r \max_{j \in M_r} \left( \sum_{i=1}^{|W|+1} x_{ij} d_i \right), \qquad \forall r \tag{20}$$

$$\text{MIP cuts} \tag{21}$$

The master problem incorporates a number of the constraints in the global MIP model. It does not represent the start-times of jobs nor does it fully represent the capacity of the trades. As is common in Benders decomposition, the master problem includes a relaxation of the sub-problems (Constraints 20) and Benders cuts (Constraints 21).

*The Sub-problem Relaxation* Defining the area of job $j$ as the area of a rectangle with height $c_{jr}$ and width $p_{jr}$, Constraint (20) is the relaxation of the capacity of a trade, expressing a limit on the area of jobs that can be executed. The limit is defined using the area bounded by the capacity of the trade and the time interval $[0, M]$ where $M$ is the maximum due date assigned to the jobs on the trade. This relaxation is due to Hooker (2005, 2007).

We tighten the relaxation of sub-problems in the Benders-MIP-T approach by enforcing an analogous limit on multiple intervals: $[0, st_w]$ for each wave $w$. For each interval, the sum of the areas of the jobs whose assigned due date is less than or equal to the end-time of the interval must be less than or equal to the available area. This relaxation is a special case of the interval relaxation due to Hooker (2005, 2007). Formally, the tighter relaxation replaces Constraint (20) with:

$$\sum_{j \in M_r} c_{jr} p_{jr} ((\sum_{i=1}^{|W|+1} x_{ij} d_i) \leq st_w) \leq st_w C_r, \quad \forall r, \; \forall w \tag{22}$$

where $((\sum_{i=1}^{|W|+1} x_{ij} d_i) \leq st_w)$ is a logical inequality evaluating to 1 if and only if the assigned due date to job $j$ is less than or equal to $st_w$.

*The Benders Cuts* Before defining the cut formally, we demonstrate the intuition with an example. Consider a due date set, $D = \{14, 17, 20, 35\}$, and, for a given trade with five jobs, the current master solution: $x_{21} = 1, x_{12} = 1, x_{43} = 1, x_{14} = 1$, and $x_{15} = 1$. Job 1 is assigned to the second due date, 17, job 2 has the first due date, 14, and so on. If the current solution is infeasible due to the resource capacity of the trade, then we know that at least one of the jobs must have a later due date than it has in the current master solution. We can, therefore, constrain the sum of the consecutive $x_{ij}$ up to and including the ones currently assigned to 1 to be one less than the number of jobs. In our example, the cut would be:

$$(x_{11} + x_{21}) + (x_{12}) +$$
$$(x_{13} + x_{23} + x_{33} + x_{43}) + (x_{14}) + (x_{15}) \leq 5 - 1$$

These variables represent the possible due dates less than or equal to those currently assigned for all jobs. By constraining these variables to be at most one less than the number of jobs, at least one job must be assigned a later due date.

Formally, assume that in iteration $h$, the solution of the DAMP assigns a set, $Q$, of due dates to the jobs on trade $r$. Assume further that there is no feasible solution on trade $r$ with the assignments in $Q$. The cut after iteration $h$ is:

$$\sum_{j \in M_r} \sum_{i \in I_{jr}^h} x_{ij} \leq |M_r| - 1, \quad \forall r \tag{23}$$

where for each job $j$ on trade $r$, $I_{jr}^h = \{i' | i' \leq i, \text{ and } x_{ij}^h = 1\}$ is the set of due date indices less than or equal to the due date index assigned to job $j$ in iteration $h$ and $|M_r|$ is the number of jobs on trade $r$. The validity of this cut is proved in Section 3.2.6.

**The Due-Date Assignment Master Problem: CP Model** We also formulate the DAMP using CP. Let $\mathcal{D}_j$ be the variable corresponding to the due date for job $j$ similar to the global CP model.

Maximize Objective (1)

Subject to:

Constraints (3) to (6), (12), (14), (15), (18)

$$\sum_{j \in M_r} c_{jr} p_{jr} \leq C_r \max_{j \in M_r}(\mathcal{D}_j), \qquad \forall r \qquad (24)$$

CP cuts (25)

The master problem modeled using CP includes several constraints of the global CP model. Constraint (24) represents the relaxation of repair capacity limit of the trades guaranteeing that the sum of processing areas for the set of jobs on the same trade does not exceed the maximum available area.

A tighter relaxation in a CP-based DAMP replaces (24) with the following defining the Benders-CP-T approach where the logical inequality ($\mathcal{D}_j \leq st_w$) evaluates to 1 if and only if the due date of job $j$, $\mathcal{D}_j$, is less than or equal to the start time of wave $w$, $st_w$.

$$\sum_{j \in M_r} c_{jr} p_{jr}(\mathcal{D}_j \leq st_w) \leq st_w C_r, \quad \forall r, \ \forall w$$

The CP cut is based on the same reasoning as the MIP cuts. If the assigned set of due dates to the jobs on trade $r$ is not a feasible solution for the SP, the cut will guarantee that in the next iteration at least one of the assigned due dates will have a greater value. Formally, the cut is:

$$\bigvee \mathcal{D}_j > \mathcal{D}_j^h, \qquad \forall j \in M_r \qquad (26)$$

where $\mathcal{D}_j^h$ is the due date assigned to job $j$ in iteration $h$, $\bigvee$ represents the logical-or constraints and $M_r$ is the set of jobs on trade $r$.

**Repair Scheduling Sub-problem** Given a set of due dates assigned to the jobs on a trade, the goal of the repair scheduling sub-problem (RSSP) is to assign start-times to the jobs to satisfy the due dates and the trade capacity. We use a CP formulation where the RSSP for each trade is modeled by the cumulative constraint.

$$\text{cumulative}([st_{jr}|j \in M_r], [p_{jr}|j \in M_r], [c_{jr}|j \in M_r], C_r), \qquad \forall r$$

$$0 \leq st_{jr} \leq \mathcal{D}_j^h - p_{jr}, \qquad \forall j, \ \forall r \qquad (27)$$

Recall that $[st_{jr}|j \in M_r]$ is the tuple of the start-time variables of the jobs on trade $r$, $\mathcal{D}_j^h$ is the value assigned to the due date for job $j$ in master problem in iteration $h$. The parameters $p_{jr}, c_{jr}, C_r$ are as defined in Table 1. Constraint (27) enforces the time windows similar to constraint (17). It is worth mentioning that in the RSSP, the due date of job $j$, $\mathcal{D}_j^h$, is a value; however, it is a decision variable, $\mathcal{D}_j$, in the global CP model.

Since CP approaches are shown to be significantly more efficient than MIP for simple scheduling problems with resource capacity constraints (Hooker & Ottosson, 2003; Hooker, 2005, 2007), we do not experiment with MIP formulations of the sub-problems.

To implement the master problems in Benders-MIP and Benders-MIP-T, we use IBM ILOG CPLEX 12.3 solver; while Benders-CP and Benders-CP-T master problems and the RSSP are implemented in IBM ILOG CP Optimizer 12.3. The details on the implementation of the global constraints are similar to Section 3.2.2.

### 3.2.4 A DISPATCHING HEURISTIC

Since the static problem is NP-hard, solving it to optimality may be prohibitively expensive. We therefore investigate a heuristic approach, inspired by the Apparent Tardiness Cost (ATC) heuristic, a composite dispatching rule that is typically applied to single machine scheduling problem with the sum of weighted tardiness objective (Pinedo, 2005). The heuristic computes a ranking index for each job and sorts the jobs in ascending order of the index. The heuristic then iterates through the jobs, scheduling each job at its earliest available time. The ranking index we use is as follows:

$$I_j = ST(k_j) \exp(-\frac{FN_j}{FC_j}), \qquad \forall j$$

If we let $k_j$ denote the type of aircraft $j$, then $ST(k_j)$ is the start-time of the first wave that requires an aircraft of type $k_j$. $FN_j$ is the fraction of the total number of aircraft of type $k_j$ required by the first wave that requires $k_j$, and $FC_j$ is the maximum proportion of the capacity needed by job $j$ over all its required trades, as follows.

$$FC_j = \max_r(\frac{p_{jr}c_{jr}}{ST(k_j)C_r})$$

Intuitively, the earlier the start-time of the first relevant wave, the higher proportion of aircraft required by that wave, and the lower the proportion of capacity required before the wave, then the sooner the job will be scheduled. The exponential function is used to place more weight on the start-time.

In preliminary experiments, three other dispatching heuristics were investigated, with the chosen heuristic performing best. The first two heuristics rank the jobs with slightly different ranking indices equal to $I_j = ST(k_j) \times \frac{\max(p_{jr})}{FN_j}$ and $I_j = ST(k_j) \times \frac{\max_r(p_{jr}c_{jr})}{FN_j}$, respectively. The third heuristic is a two-stage approach based on a decomposition. The first stage finds the number of each aircraft type assigned to each wave and the second stage schedules the jobs in increasing order of $\max_{j,r}(p_{jr}c_{jr})$ considering the values determined in the first stage as upper bounds on the number of jobs required before each wave. Our preliminary experiments demonstrated that our chosen dispatching rule results in on average of 6% higher wave coverage compared to the first two heuristics and in the same coverage as the third heuristic while having the advantage of being easy to understand and implement.

### 3.2.5 HYBRID HEURISTIC-COMPLETE APPROACHES

A hybrid heuristic-complete approach in which the heuristic solution provides a lower bound for the maximization objective (Equation 1) may improve the performance of the complete approaches. Therefore, a simple hybrid first runs the dispatching heuristic and then uses the objective value as a starting lower bound for the complete approaches. Assume that the heuristic finds a solution, $S$, with $f(S)$ as the number of aircraft assigned to waves. Any of the complete approaches can now be modified by adding the following constraint:

$$\sum_{w=1}^{|W|} \sum_{k=1}^{|K|} Z_{kw} \geq f(S)$$

For LBBD variations, the above constraint is added to the master problem.

### 3.2.6 Theoretical Results

To guarantee the finite convergence of a LBBD model to a globally optimal solution, the Benders cuts must be valid and the master decision variables must have finite domains. A Benders cut is valid in a given iteration, $h$, if and only if (1) it excludes the current globally infeasible assignment in the master problem without (2) removing any globally optimal assignments (Chu & Xia, 2004). The former guarantees the finite convergence and the latter guarantees the optimality. As the decision variables in DAMP have a finite domain, it is sufficient to prove the satisfaction of the two conditions.

**Theorem 2.** *Cut (23) is valid.*

*Proof.* For condition (1), for the sub-problem in iteration $h$ on trade $r$, by definition:

$$\sum_{j \in M_r} \sum_{i \in I_{jr}^h} x_{ij} = |M_r|$$

Consequently, cut (23) excludes the current assignment of master problem.

For condition (2), consider a global optimal solution $S$ that does not satisfy cut (23) as generated in iteration $h$. As the cut states that at least one job must have a greater due date than it had in $h$, to violate the cut, all jobs in $S$ must have equal or lesser due dates than they had in iteration $h$. However, because the sub-problem was infeasible in iteration $h$, any sub-problem with only equal or lesser due dates must also be infeasible as the available capacity on the trade is the same or less. Therefore, $S$ must be infeasible and we contradict the assumption that $S$ is globally optimal.

Therefore, the cut is valid. □

An analogous argument holds for cut (26).

### 3.3 Rescheduling Strategies

The dynamic repair shop problem over the long horizon can be viewed as static scheduling sub-problems over successive time periods. Let's assume that we start repairing the failed aircraft and assigning them to the waves based on the computed schedule at time 0. A wave might start while a repair is under way in the repair shop. If some aircraft fails the pre-flight check, it goes to the repair shop. Each failed aircraft requires a set of independent repair activities with known processing times and resource requirements. At the repair shop, some of the previously failed aircraft might be already repaired, some might be under repair, and others might be awaiting repair. Once the failed aircraft enters the repair shop, we have a new static repair scheduling sub-problem where the set of existing jobs ($J$), the number of aircraft not in the repair shop for each aircraft type ($\eta_k$), and the failure rates of the aircraft ($\lambda_n$) are updated. The set of existing jobs includes the recently failed aircraft and the previously failed aircraft whose repairs are still under way or are not yet started. The new static sub-problem has an added constraint, namely that the repairs currently under way cannot be disrupted.

We connect the static sub-problems using three different policies denoted as $P_{ij}$ where $i$ and $j$ define the length of scheduling horizon and the frequency of rescheduling in number of waves, respectively. In all three policies, we schedule the repair activities, observe the aircraft failures, and respond to failures by rescheduling the repair activities.

The three policies discussed here are as follows:

- $P_{11}$: In Figure 5, we show that $P_{11}$ schedules one wave at a time ($i = 1$) and reschedules after each wave ($j = 1$). $P_{11}$ is a myopic policy aiming at providing the next first wave with the highest possible coverage.

- $P_{31}$: In contrast to $P_{11}$, for $P_{31}$ (Figure 5), the scheduling horizon is three waves but rescheduling is still done after each wave. $P_{31}$ with a longer scheduling horizon than $P_{11}$ trades-off the coverage among the next three waves. It is worth mentioning that we have chosen three as the length of the scheduling horizon because three waves are usually scheduled daily based on the real data (Safaei et al., 2011).

- $P_{33}$: This policy has a scheduling horizon with a length of three waves and reschedules after every third wave (Figure 5). $P_{33}$ might trade-off the lower coverage of the next first wave for higher coverages of the second and the third waves; however, it has a lower frequency of rescheduling.

### 3.4 Modeling the Aircraft Failures

To model the dynamic events, we simulate the aircraft failures in pre- and post-flight checks. Every aircraft either passes or fails each check. If the aircraft fails, a new set of repair activities with known processing times and resource requirements is added to the repair shop. If the aircraft passes, it flies the wave if required. As mentioned in Section 2.1, after repair the failure rate of the aircraft returns to what it was before the failure and it increases by $\gamma$ percent each time it flies a wave due to deterioration. If $\lambda_n$ is the initial failure rate of the aircraft $n \in N$, its failure rate after flying $w$ waves without failure will be $\lambda_n(1 + 0.01 \times \gamma)^w$.

## 4. Computational Experiments

In this section, we present two separate empirical studies. The first study compares the scheduling techniques experimentally and presents insights into each algorithm's performance through a deeper analysis of the results. The second study investigates the impact of using different scheduling techniques and rescheduling policies on the observed wave coverage.

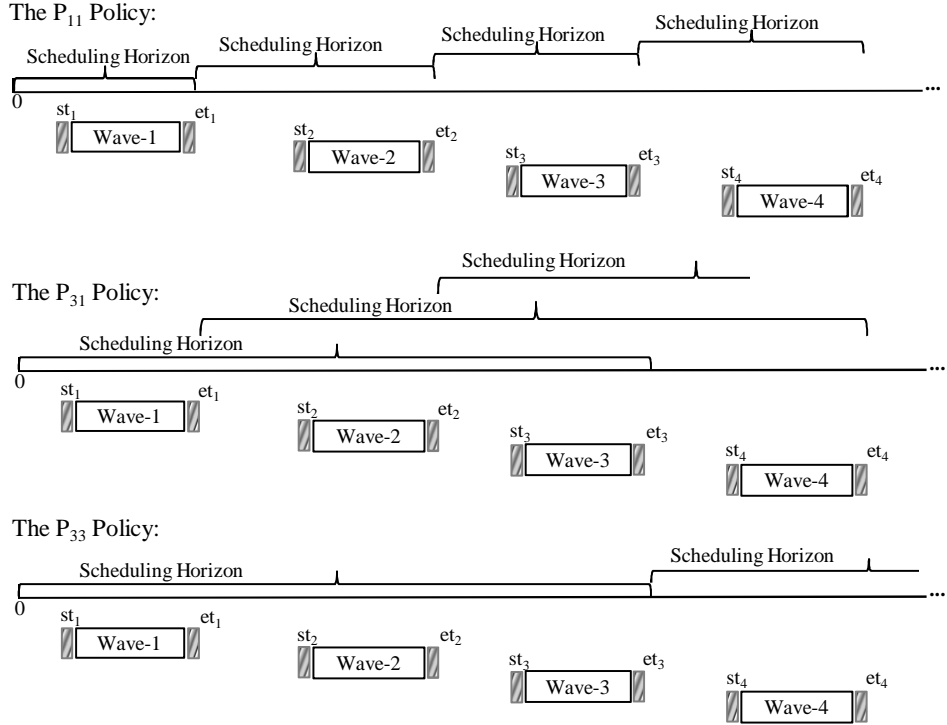

Figure 5: The rescheduling policies.

## 4.1 Experimental Results on Scheduling Techniques

This sub-section describes the experiment comparing different solution techniques for scheduling the static repair shop.

### 4.1.1 Experimental Setup

The problem instances have 10 to 30 aircraft (in steps of 1), 3 or 4 trades, and 3 or 4 waves. Five instances for each combination of parameters are generated, resulting in 420 instances (21 total aircraft counts by 2 trades counts by 2 waves counts by 5 instances).

*Aircraft*: The number of aircraft types is equal to $\frac{|N|}{5}$, where $|N|$ is the number of aircraft. The aircraft are randomly assigned to different types with uniform probability. The number of aircraft of type $k$ is $|I_k|$. The failure rate for each aircraft is randomly chosen from the uniform distribution $[0, 0.5]$. The failure rate for aircraft of type $k$, $\bar{\lambda}_k$, is the mean failure rate over all aircraft of type $k$. The functions used to represent aircraft $n$ probability of failures in pre- and post-flight checks, respectively, are $f^{pre}(\lambda_n) = (1 - e^{-\lambda_n})$ and $f^{post}(\lambda_n) = (1 - e^{-3\lambda_n})$. It is worth mentioning that the conditions of a reliability function in the extreme values of failure rates hold true for the functions used. If the failure rate goes to 0, the probability of failure equals 0, and if the failure rate goes to $\infty$, the probability of failure equals 1.

*Waves*: The plane requirement of each aircraft type for each wave is randomly generated from the integer uniform distribution $[1, |I_k|]$. The length of each wave is drawn with uniform probability from $[3, 5]$. To make an instance loose enough to permit feasible solutions yet tight enough to be challenging, a lower bound on the length of the scheduling horizon ($H$) is needed. The sum of the processing areas of the jobs in each trade, $r$, divided by the trade capacity is denoted by $S_r$. $LB = \max_r(S_r)$ is a lower bound on the time required to schedule all jobs and we use $H = 1.2 \times LB$. The end-time for each wave, $et_w$, is generated as $et_{|W|} = H - rand[0, 3]$ for the final wave, $|W|$, and $et_w = st_{w+1} - rand[0, 3]$ for $w < |W|$.

*Trades*: The capacity limit for each trade is set at $C_r = 10$.

*Repair Jobs*: Eighty percent of the aircraft are in the repair shop at the beginning, resulting in $|J| = 0.8|N|$ repair jobs. The jobs are randomly assigned to the trades with replacement such that the number of jobs per trade is equal to $|J|/2$. Each job requires at least one trade and some require more than one trade. The capacity of trade $r$ used by job $j$, $c_{jr}$, is drawn from $[1, 10]$ while the processing time, $p_{jr}$, is drawn from $[r, 10r]$: jobs on trades with lower indices have shorter processing times than those on trades with higher indices.

Though our problem instances are generated randomly, the setting of our experiment includes three numerical examples of Safaei et al. (2011) which are based on the real data. Furthermore, our setting consists of more instances and results in problem instances which are one and a half times bigger than the examples used in the literature (Safaei et al., 2011) where the number of aircraft is 10, 15, or 20; the number of waves is 3 or 4; and the number of trades and aircraft types is equal to 3 and 2.

All experiments were run with a 7200-second time limit on an AMD 270 CPU with 1 MB cache per core, 4 GB of main memory, running Red Hat Enterprise Linux 4.

### 4.1.2 Experimental Results

Figure 6 shows scatter-plots of run-times of the six complete approaches. Both axes are log-scale, and the points below the line $y = x$ indicate a lower run-time for the algorithm on the y-axis. The numbers in the boxes indicate the number of points below or above the line. Run-times are counted as equal if they differ less than 10%.

The graphs indicate a benefit for MIP over CP, for Benders-CP over MIP, for Benders-MIP over Benders-CP and MIP, for Benders-MIP-T over Benders-MIP, and for Benders-CP-T over Benders-CP. Table 2 presents further data, sorted in descending percentage of problems solved to optimality, for all algorithms. The changes in the results compared to the previous work (Aramon Bajestani & Beck, 2011b) are due to improved scheduling models, different solvers, and different test problems. The scheduling algorithms are enhanced by using a smaller value for $B$, the resource capacity constraint is enforced in a shorter interval, i.e., $[0, st_{|W|}]$, and more efficient formulations for Constraints (5), (6), and (20) are used.

The mean run-time of the MIP model given by Safaei et al. (2011) over eight scenarios with 10 aircraft and 3 waves is 294.75 seconds. However, our proposed MIP model has the run-time of 2.64 seconds on average over ten of the instances with the same number of aircraft and waves, indicating that it is significantly faster than Safaei et al.'s model.

**MIP vs. CP**   The MIP approach has a clear superiority over the CP, achieving a lower run-time on 89% of the problem instances. The CP model outperforms the MIP only on 5% of the instances where it can solve to optimality within the time limit. A further investigation of the results shows that the mean quality of CP solution is 0.27% from the best found solution across all the algorithms. Therefore, the poor performance of CP is due to its weakness in proving the optimality.

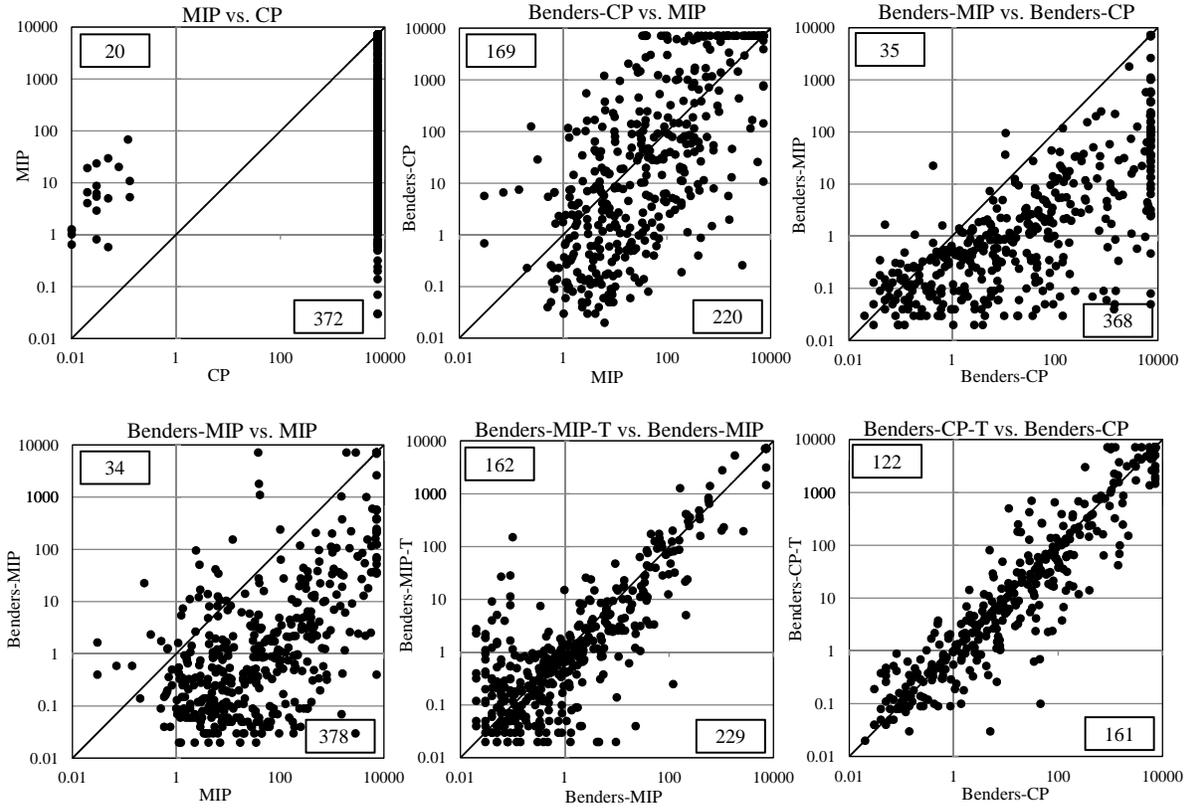

Figure 6: Run-times (seconds) of the six complete models.

| Method | Mean Time (s) | Iter. | % MP | % SP | % Solved to optimality |
|---|---|---|---|---|---|
| Benders-MIP-T-Hybrid | 211.98 | 66.63 (8.0) | 51.39 (55.05) | 48.61 (44.95) | 98.10 |
| Benders-MIP-T | 213.12 | 66.44 (8.0) | 51.84 (53.98) | 48.16 (46.02) | 97.86 |
| Benders-MIP | 227.94 | 64.66 (8.0) | 61.75 (67.44) | 38.25 (32.56) | 97.62 |
| MIP | 837.04 | - | - | - | 93.57 |
| MIP-Hybrid | 924.30 | - | - | - | 91.19 |
| Benders-CP | 1373.16 | 75.72 (15.5) | 84.30 (96.96) | 15.70 (3.04) | 85.24 |
| Benders-CP-T | 1356.70 | 66.42 (10.0) | 85.36 (97.36) | 14.64 (2.64) | 85.00 |
| Dispatching Rule | $\approx 0$ | - | - | - | 9.76 |
| CP | 6857.14 | - | - | - | 4.76 |

Table 2: The mean run-time, the mean (the median) number of master problem iterations, the mean (median) percentage of run-time spent solving the master problem and the sub-problems, and the percentage of problems solved to optimality for all approaches.

**Benders-CP vs. MIP**  The Benders-CP approach does better than MIP in terms of run-time on 52% of the instances while performing worse on 40%. However, Table 2 favors MIP in terms of the overall performance, smaller mean run-time, mainly because more instances are solved to optimality within the time limit.

**Benders-MIP vs. Benders-CP**  The Benders-MIP approach achieves a better run-time than Benders-CP on 88% of the test problems, performing worse on about 8%. The branching heuristics for Benders-CP often lead to an initial feasible master solution with tighter due dates than the initial master solution in Benders-MIP. The tighter, globally infeasible initial solution means that the CP-based master problem model requires more iterations to find a globally feasible solution.

**Benders-MIP vs. MIP**  The Benders-MIP approach achieves a better run-time than MIP on 90% of the test problems and a worse run-time on 8%, achieving a lower mean run-time and solving a higher proportion of the problem instances. When the time horizon is short, the MIP approach is faster, however, with longer horizons and more jobs, the number of constraints and variables grows, substantially reducing its performance.

**Benders-MIP-T vs. Benders-MIP**  The tighter relaxation in Benders-MIP-T slightly speeds up LBBD: Benders-MIP-T has a better run-time than Benders-MIP on 55% of problems instances and worse on 39%. The mean run-time decreases by 6% and the tighter relaxation solves one more instance to optimality. Tightening the relaxation of sub-problems increases the mean number of iterations in spite of our expectation. A closer look to the results shows that the mean number of iterations of the instances solved to optimality by both approaches (97.38% of the instances) decreases: 39.15 and 41.24 for Benders-MIP-T and Benders-MIP, respectively. However, the mean number of iterations of the instances timed out by both approaches (1.9% of the instances) increases from 1051.38 for Benders-MIP to 1259.88 for Benders-MIP-T. Therefore, the increase in the number of iterations results from the timed-out instances which does not support that the tighter relaxation requires more iterations to optimality.

Furthermore, the percentage of time solving the master problem decreases compared to Benders-MIP, while the sub-problem percentage run-time increases. This latter observation is because the sub-problems for Benders-MIP that can be quickly proved insoluble by the initial propagation of CP sub-problem model, violate the tighter relaxation in the Benders-MIP-T master problem. Therefore, in the tighter model, the sub-problem solver is not called on these "easy" sub-problems, increasing the percentage run-time spend on sub-problems.

**Benders-CP-T vs. Benders-CP**  The tighter relaxation in CP-based master problem results in a slightly lower mean run-time; however, as shown in Figure 6, their performance comparison is more even.

**Incomplete and Hybrid Approaches**  The dispatching heuristic is fast, finding a feasible solution to all problems. However, it finds (but, of course, does not prove) an optimal solution in only 9.76% of the instances and Benders-MIP-T finds and proves optimality for these instances in 0.99 seconds on average. It seems that the heuristic can find the optimal solution only when the problem instance is relatively easy. The mean quality of the heuristic solution is 16% from optimal. In industries with expensive assets, such a reduction in solution quality can translate to costly under-use of a valuable resource (e.g., a fighter aircraft costs in the vicinity of 100 million dollars). However, finding a feasible solution almost instantaneously for large problem instances makes the heuristic approach compelling in situations where the long run-time might delay carrying out the waves. For example, if a wave starts within a very short time, flying the wave with lower coverage (achieved by the dispatching heuristic) is better than not carrying it out because of the long solving time of the complete approaches.

To evaluate the effect of combining the dispatching heuristic with the complete approaches, we examine using the hybrid heuristic-complete approach. A smaller feasible set is the direct consequence of defining a bound on the cost function. As the MIP model searches the feasible set, while LBBD methods explore the infeasible space, one intuition is that the MIP model should benefit more from using the heuristic solution. However, solving the master problem in LBBD requires searching in a relaxed feasible space and therefore the heuristic starting solution may also speed solving.

Table 2 shows a very marginal benefit for bounding the Benders-MIP-T approaches with the dispatching heuristic solution. Bootstrap paired-$t$ tests (Cohen, 1995) also indicate that there is no significant difference in mean run-time at $p \leq 0.01$ for either hybrid.

**Scalability**   Figure 7 shows our results as the number of aircraft per wave increases. We aggregate results by truncating $\frac{|N|}{|W|}$ and using the instances with three waves and both three and four trades. Note that each point represents 30 problem instances except $x = 3$ which has only 20 problems instances. We omitted $x = 10$ as we only have 10 problem instances for that point. The $y$-axis is log-scale.

The results show that the LBBD variations outperform the other techniques across all ratios.

**Summary**   The following observations on the performance of scheduling techniques are supported by this empirical study.

- The LBBD approach combining mixed integer programming and constraint programming outperforms the mixed integer programming model. The mean run-time of Benders-MIP is almost 4 times lower than MIP. Furthermore, defining the time ratio for a given instance as the MIP run-time divided by Benders-MIP run-time, the Benders-MIP is almost 32 times faster than MIP, on average, with a geometric mean time ratio of 31.56. The time ratios range is $[0.01, 94132.67]$ with a median of 38.32.

- A tighter relaxation slightly speeds up LBBD. Benders-MIP-T and Benders-CP-T, both, have a run-time about 1.2 times faster than Benders-MIP and Benders-CP, respectively.

- A dispatching heuristic can provide the optimal solution for the easy problem instances. However, the mean percent relative error of heuristic is almost 16% overall, indicating that the dispatching rule by itself is not effective enough for industries with high equipment cost.

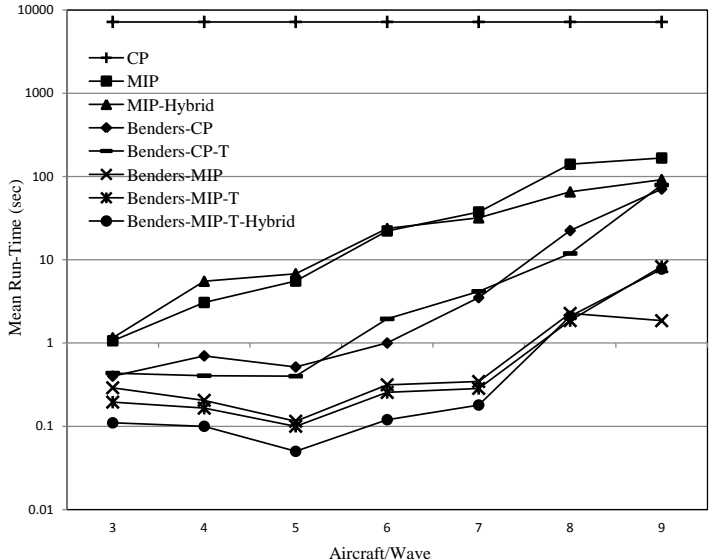

Figure 7: Mean run-time vs. number of aircraft per wave ($|W| = 3$).

- The simple hybridization of the complete approaches with the dispatching heuristic does not result in a statistically significant difference in run-time.

## 4.2 Experimental Results on Rescheduling Strategies

This sub-section describes experiment investigating the impact of applying different scheduling techniques and rescheduling policies in a dynamic repair shop.

### 4.2.1 EXPERIMENTAL SETUP

For our problem instances, the number of aircraft, the number of trades, and the total number of waves are set to $\{10, 15, 20, 25, 30\}$, $\{4\}$, and $\{30\}$ respectively. Each combination has 5 instances for a total of 25 instances. Each instance is simulated 20 times. The parameters of the problem instances are generated as in Section 4.1 with the following modifications:

*Aircraft*: The failure rate of an aircraft is increased by $\gamma = 5$ percent each time it is used.

*Repair Jobs*: Repair jobs that entered repair shop after time 0 are randomly assigned to the trades. The probability of assigning a job to each trade is considered as 0.5.

*Waves*: The start-time of each wave is generated as $st_1 = rand[\frac{H}{3}, \frac{H}{2}]$ for the first wave, and $st_w = et_{w-1} + rand(0, 40)$ for $1 < w \leq 30$. As mentioned earlier the total number of waves is 30. The value of $H$ is calculated as in Section 4.1.

*Dynamic events*: To simulate an aircraft failure, we generate a random value from the uniform distribution $[0, 1]$ for each aircraft at each check. If the random value is less than the aircraft's probability of failure, the aircraft fails; otherwise, it passes. The aircraft's probability of failure in pre- and post-flight checks are calculated using $(1 - e^{-\lambda_n})$ and $(1 - e^{-3\lambda_n})$, respectively. Recall that, $\lambda_n$ is the failure rate of aircraft, $n \in N$, which increases by $\gamma = 5$ percent each time the aircraft flies a wave. Note that passing the pre-flight check of a wave does not necessarily mean that the aircraft flies the wave. If the number of available aircraft is more than the requirements, the aircraft that fly are randomly selected from those that passed the pre-flight check to meet the requirements. Our latter assumption implies that all aircraft ready at the beginning of a wave are checked regardless of the wave requirements. Since we have assumed that the pre- and post-flight checks have negligible cost, our assumption is reasonable to discover the potential aircraft failures sooner which is likely to increase their availability for the subsequent waves.

We experiment with three techniques including MIP, Benders-MIP-T, and the dispatching rule discussed in Section 3.2. The time-limit to schedule the repair activities at each decision time point is 600 seconds. We execute the best feasible schedule found before the time-limit if an algorithm times out. In the case that Benders-MIP-T times out, the schedule created by the dispatching heuristic is executed as Benders-MIP-T does not create a feasible schedule when it times-out.

As in the static problem, the scheduling uses IBM ILOG CPLEX 12.3 and IBM ILOG CP Optimizer 12.3. The simulation is implemented in C++.

### 4.2.2 Experimental Results

In this section, we discuss our results to compare the performance of different scheduling and rescheduling techniques on the availability of the aircraft in the long run. We further investigate the effect of modeling the aircraft failures using the expected coverage.

Figures 8, 9, and 10 illustrate the mean observed coverage up to flight $w \in \{1, 2, ..., 28\}$ for different scheduling and rescheduling techniques. Denoting $\nu_{wpl}$ as the coverage of flight $w$ in the $l$-th simulation of instance $p$ for a given policy, $O_{wpl} = \frac{\sum_{i=1}^{w} \nu_{ipl}}{w}$ represents the mean observed coverage up to flight $w$ in instance $p$ and in simulation $l$. The mean observed coverage up to flight $w$, shown in the figures, is calculated as $\mathcal{O}_w = \frac{\sum_{p=1}^{P} \sum_{l=1}^{L} O_{wpl}}{PL}$, where $P$ and $L$ are the number of instances and simulations, respectively. Table 3 shows the mean observed coverage up to flight 28, i.e., $\mathcal{O}_{28}$ and the variance of the observed coverage up to flight 28 for all scheduling techniques and rescheduling policies. As illustrated, Benders-MIP-T using $P_{31}$ achieves at least a 10% higher mean coverage than any other combinations of the scheduling and rescheduling techniques and has the lowest variance.

**The impact of the scheduling algorithms** A complete technique is anticipated to achieve higher flight coverage because it takes the expected probabilistic information into account when creating a repair schedule, while the dispatching heuristic does not have this property. As shown in Table 3, Benders-MIP-T as a complete technique results in higher mean observed coverage over all policies when compared to the dispatching heuristic. However, MIP, incorporating the mean of known information on uncertainty into scheduling the repair activities, results in flights with lower coverage than the dispatching heuristic over two of the rescheduling policies, $P_{11}$ and $P_{33}$. To understand the MIP performance, we make two conjectures.

| Method | $\mathcal{O}_{28}[\text{var}(.)]$ | | | $\rho(.)$ | | |
|---|---|---|---|---|---|---|
| | $P_{11}$ | $P_{31}$ | $P_{33}$ | $P_{11}$ | $P_{31}$ | $P_{33}$ |
| Benders-MIP-T | 0.67 [0.03] | 0.77 [0.01] | 0.70 [0.01] | 45 | 56 | 48 |
| MIP | 0.52 [0.03] | 0.64 [0.03] | 0.60 [0.02] | 34 | 46 | 38 |
| Dispatching heuristic | 0.61 [0.04] | 0.61 [0.04] | 0.63 [0.03] | 42 | 44 | 47 |

Table 3: The mean (variance) of observed coverage up to flight 28 ($\mathcal{O}_{28}[\text{var}(.)]$) and the mean percentage of available aircraft for the first flight ($\rho$).

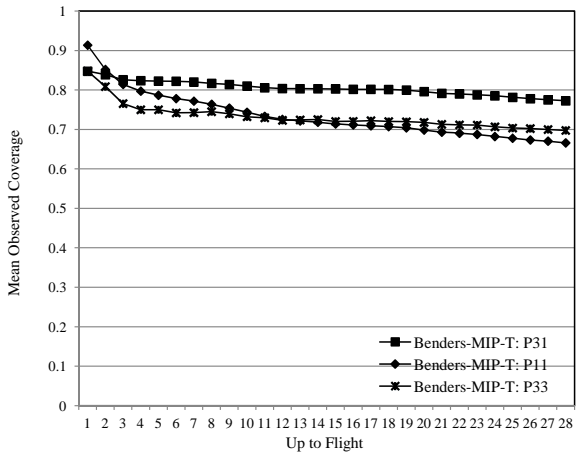

Figure 8: Mean observed coverage for three different policies using Benders-MIP-T.

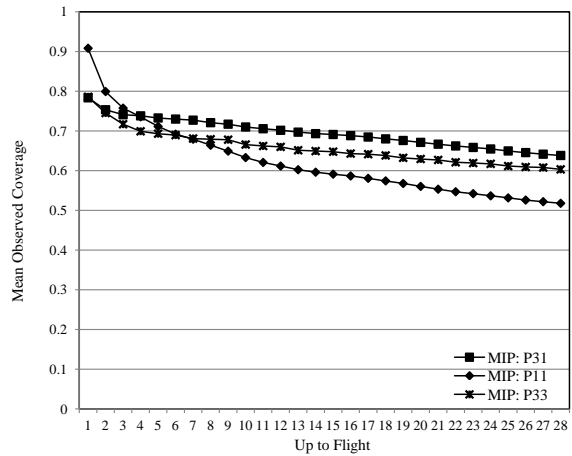

Figure 9: Mean observed coverage for three different policies using MIP.

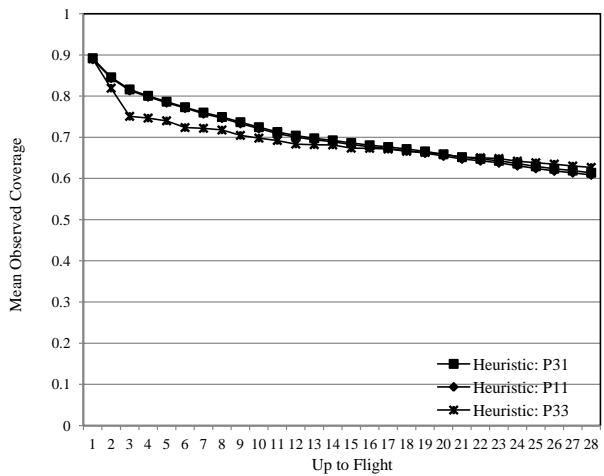

Figure 10: Mean observed coverage for three different policies using the dispatching heuristic.

Our first conjecture is that the poor performance of MIP algorithm is because it frequently times out on most of the static sub-problems and the best found feasible solution or the dispatching heuristic is then used to create the repair schedule. However, our results do not support the conjecture. MIP algorithm only times out on 13% of the scheduling sub-problems and a feasible solution is found in each, implying that the dispatching heuristic is never used to find a repair schedule.

Our second conjecture is that the low coverage achieved by MIP can be attributed to its different way of scheduling the repair activities compared to the other two scheduling techniques. A deeper look into the schedules of the static sub-problems shows that the dispatching heuristic and Benders-MIP-T schedule the repair activities at the earliest possible time; however, MIP usually does not. Repairing the aircraft earlier makes more aircraft available which intuitively increases the coverage in the long run, even though the number of pre-flight checks that aircraft go through increases in expectation. The quick adjustment of the schedule makes some of the failed planes again available before the start-time of the next flight. To investigate the impact of making the aircraft available earlier using a given scheduling technique, we define the mean percentage of available aircraft for the first flight as $\rho(P_{ij}) = \frac{\sum_{p=1}^{P} \sum_{l=1}^{L} \sum_{k=1}^{S} \rho_{kpl}}{PLS}$ where $\rho_{kpl}$ and $S$ denote the percentage of aircraft available at the beginning of the first flight of the $k$-th static sub-problem in the $l$-th simulation of instance $p$ and the number of static sub-problems in $P_{ij}$ policy, respectively. For example, in $P_{31}$ policy for a given $p$ and $l$, the first static sub-problem includes flights 1, 2, and 3. Then, $\rho_{1pl}$ is the number of aircraft available for flight 1 divided by the total number of aircraft. The second static sub-problem schedules for flights 2, 3, and 4. Therefore, $\rho_{2pl}$ is equal to the number of aircraft available for flight 2 divided by the total number of aircraft. We follow the same procedure to find $\rho_{kpl}$ for all 28 static sub-problems in $P_{31}$ policy. We can find $\rho(P_{11})$ and $\rho(P_{33})$ using the same argument considering that the number of static sub-problems are 30 and 10, respectively.

Comparing any pair of the scheduling and the rescheduling techniques in Table 3, there is a positive relationship between making the aircraft available earlier and the wave coverage in the long term which supports our conjecture: if the mean percentage of available aircraft for the first flight ($\rho$) increases, the mean observed coverage in the long rum ($\mathcal{O}_{28}$) also increases.

**The impact of rescheduling policies** As illustrated in Figures 8 and 9, the $P_{11}$ policy with either Benders-MIP-T or MIP in the short-term (i.e., for the first three flights) outperforms the other two policies. However, the $P_{31}$ policy then leads to consistently higher coverage because it schedules over a longer horizon and adjusts the schedule as soon as aircraft failures occur. Although $P_{31}$ with the dispatching heuristic also responds quickly to the aircraft failures, it does not incorporate the length of the scheduling horizon into the ranking index for repair activities and always repairs the aircraft for the earliest possible time, resulting in flights with the same coverage as $P_{11}$.

Figure 11 displays the cumulative percentage of the flights with a coverage less than or equal to $\omega$ for Benders-MIP-T, the dispatching heuristic and MIP where $\omega$ denotes the values on the x-axis. The best performing approach will have fewer flights with a low coverage and more flights with a high coverage. Therefore, its curve will be closer to the lower right-hand corner. As illustrated, Benders-MIP-T using $P_{31}$ performs better than any other combination. The $P_{31}$ rescheduling policy is computationally more expensive than the other two policies, its run-time per one static sub-problem, however, is small compared to the length of scheduling horizon being usually one day in the real applications (Safaei et al., 2011). The $P_{31}$ policy using Benders-MIP-T has a run-time of on average 67 seconds per one static sub-problem and of less than 249 seconds on 90% of static sub-problems.

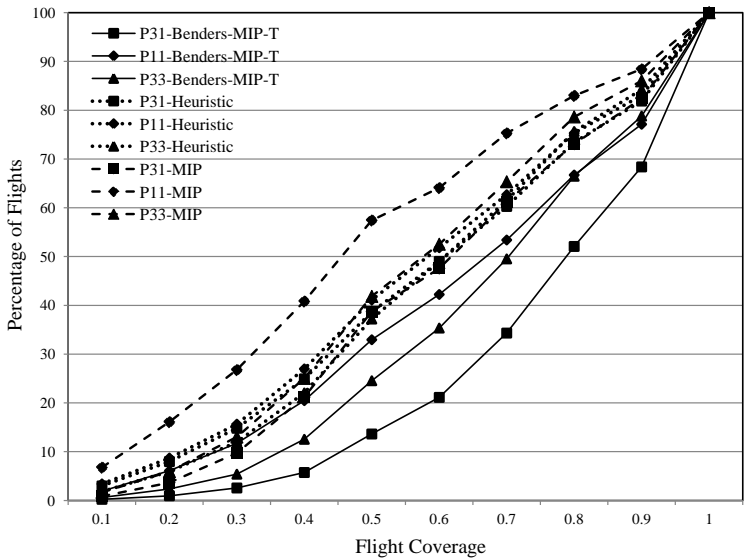

Figure 11: The percentage of flights with a coverage less than or equal to $\omega$, where $\omega$ denotes the values on the x-axis.

In summary, our analysis of the results identifies the Benders-MIP-T with $P_{31}$ (Benders-MIP-T:$P_{31}$) as the best combination of the scheduling and the rescheduling techniques providing the flights with a higher mean coverage over the long term. Furthermore, it has the lowest variance for the observed coverage compared to the other scheduling and rescheduling techniques.

**The impact of modeling the uncertainty in expectation**   Because of the random aircraft failures, the coverage achieved by any scheduling algorithm is a random variable. The ultimate goal is to construct a repair schedule which is optimal for the specific realization of the uncertainty that actually occurs. However, since the complete information on the aircraft failures is not known and the future uncertainty is dependent on the previous repair decisions, it is impossible to find a repair schedule which is ideal under any realization of uncertainty. As discussed earlier, we have modeled the aircraft failures using the expected value to find the optimal repair schedule. Since treating the uncertainty in the expectation form may be far from optimal for the actual realization of uncertainty, we perform a sensitivity analysis on the failure rates of the aircraft to investigate how the optimal repair schedule by Benders-MIP-T:$P_{31}$ is hedged against various uncertain situations.

Using the same problem instances as in Section 4.2.1, two other experiments are set up where the failure rate of each aircraft ($\lambda_n$) is increased to $\lambda_n + 0.05$ and $\lambda_n + 0.1$. Our results show that while the mean observed coverage up to flight 28 decreases to 0.69 and 0.62, the variance of observed coverage does not change, indicating that modeling the uncertainty using the expected probabilistic information is a reasonable approach.

To find a possible upper bound (tighter than 1) on the mean observed coverage up to flight $w$ under any scheduling algorithm, we define a policy called "Relaxed" which relaxes the repair capacity limit and repairs any failed aircraft after its maximum processing times over all trades. Although the Relaxed policy makes the aircraft available for the waves earlier than any other repair scheduling policy, we cannot guarantee that it results in the upper bound on the observed coverage unless all waves have the same requirements. The optimal decision might be to trade off the immediate low coverage for future higher coverage when the plane requirements for the waves are different. Applying the Relaxed scheduling policy on the same instances as in Section 4.2.1, the mean observed coverage up to flight 28, i.e. $\mathcal{O}_w$, is 24% higher than the best identified algorithm, Benders-MIP-T:$P_{31}$. More specifically, the Relaxed policy results in a mean coverage of 0.95 with the variance of 0.002.

**The impact of longer scheduling horizon vs. more frequent rescheduling** The $P_{31}$ policy changes the repair schedule after each flight and trades-off the coverage among three consecutive flights by scheduling over a longer horizon. In contrast, the $P_{11}$ policy schedules for one flight and reacts after each flight while the $P_{33}$ policy reasons over a longer term without a quick response to the dynamic events. As already shown in Figures 8 and 9, the $P_{31}$ policy with any of the complete techniques results in a higher mean coverage. The $P_{11}$ policy outperforms the $P_{33}$ for early waves, but the $P_{33}$ provides the later waves with higher coverage.

The superiority of policy $P_{31}$ indicates that both features of quick response to the dynamic events and long-term reasoning contribute to the overall performance. The contribution of each feature is significantly dependent on several parameters such as the aircraft failure rates, the plane requirements, and the repair capacity. When the failure rate is high, the probability of aircraft being diagnosed as failed in pre- and post- flight checks is higher. Therefore, the arrival rate of the aircraft to the repair shop is higher and the previously constructed schedule is more likely not to be executed as is. In such a system, frequent rescheduling is more likely to increase the coverage. When the plane requirements of the waves are widely varying and the repair capacity limit is tight, trading-off the coverage among the flights through scheduling over a longer horizon significantly contributes to the availability of the aircraft in the long term.

**Summary** The following observations on how and when the scheduling and rescheduling should be done are supported by the second empirical study:

- Solving the dynamic repair shop problem using the Benders-MIP-T scheduling technique and the $P_{31}$ rescheduling policy results into an observed coverage with higher mean and lower variance than any other combination tested.

- There is a positive relationship between making the failed aircraft available as early as possible and achieving a higher coverage in the long term.

- Since the variance of the $P_{31}$ policy with the Benders-MIP-T is not sensitive to the small changes in the aircraft failures, modeling the uncertainty with respect to the mean is a reasonable approach to balance against different uncertain scenarios.

## 5. Discussion

The experimental results demonstrate that incorporating both probabilistic and execution time reasoning into the schedule of repair activities results in a better system performance. We showed that the decomposition technique, LBBD, and the rescheduling policy $P_{31}$ result in a 10% higher mean observed coverage in the long term, increasing the utilization of the valuable resources. The decomposition technique considers the mean of known probabilistic information about uncertainty over a longer scheduling horizon and repairs the failed aircraft at the earliest time. The $P_{31}$ rescheduling policy takes advantage of up-to-date information more frequently. It is also shown that the variance of the coverage does not increase as the aircraft failures increase, supporting the core idea of our solution approach: the dynamic repair shop problem is viewed as a collection of static sub-problems where the uncertainty on aircraft failures is treated as expectation.

Optimizing with respect to the mean and considering a specific class of scheduling problems are limitations of our solution approach. We address each in detail below and discuss ideas to deal with them.

**Modeling the uncertainty**  Optimizing with respect to the expected coverage can have unfavorable consequences: the constructed repair schedule may have a remarkably poor performance for particular realizations of uncertainty that might happen in actuality (Birge & Louveaux, 1997). There are a number of other possible approaches for solving the dynamic problem. We briefly discuss each method below.

Leaving some availability slack on repair resources to make the schedule more robust and flexible (Branke & Mattfeld, 2002, 2005; Davenport, Gefflot, & Beck, 2001) is the first approach. For example, Branke and Mattfeld (2002, 2005) propose an anticipatory scheduling algorithm to predict the future job arrivals in a dynamic scheduling problem. A secondary objective, called flexibility, is included within each static sub-problem to penalize the early idleness of the machines. They experimentally show that this approach improves the system performance. Their conclusion is consistent with our observation in Section 4.2.2 on the positive relationship between repairing the aircraft earlier and achieving a higher coverage. It would be therefore interesting to adjust the MIP model such that a flexibility term is added in the objective function to quantify the value of making the repair resources available as early as possible. However, it appears that none of the existing work on such slack-based techniques uses analytical reasoning to decide the amount of slack or level of penalization of early slack that should be used for different levels of stochasticity.

Modeling each static sub-problem as a two-stage stochastic programming is the second approach (Birge & Louveaux, 1997). The first stage decision corresponds to constructing the repair schedule and occurs before aircraft failures in pre- and post-flight checks. The second stage decision, which includes the allocation of aircraft to flights, occurs after the pre-flight checks. One approach is to define $Z_{kw}^s$ as the number of aircraft of type $k$ assigned to fly in wave $w$ under scenario $s$. Each scenario $s$ represents a possible realization of aircraft failures in the horizon of the static sub-problem with probability $p(s)$. Therefore, the objective function (Equation 1) can be written as $\sum_s p(s) \sum_k \sum_w Z_{kw}^s$. The main modeling challenge is then to calculate the probability of each scenario. As already explained in Section 2.1, the uncertainty in our problem is not exogenous information and is dependent on the first-stage decisions which is hard to represent it in a closed and tractable form. Computationally, two-stage stochastic programming models are substantially more challenging than most discrete optimization problems (Dyer & Stougie, 2003) and therefore the ability to solve such models of our problem to optimality is doubtful.

The third approach is to use multi-stage dynamic programming for solving the dynamic repair shop problem (Iravani et al., 2007). The goal is to construct a repair schedule at each decision epoch, marked by the arrival of newly failed aircraft to the repair shop, such that the coverage is maximized over the long term. Using the dynamic programming framework, the state of the repair shop at each decision time point is a tuple of the aircraft failure rates, the aircraft processing times, and the aircraft resource requirements. The decision or action is to assign start-times to the failed aircraft in the repair shop. There are several challenges in modeling the problem as a classical dynamic program. First, the expected wave coverage as a result of the current state and the action taken cannot be represented in a closed form expression because of the combinatorics involved in the scheduling problem. Second, the probabilities with which the repair shop transitions to a new state at the next decision epoch as a result of the current state, the action taken, and the revealed uncertainty on the aircraft failures are not known and are hard to calculate mainly, again, due to the combinatorics of the scheduling decisions and the fact that the processing times and the resource requirements of the repair operations become known upon aircraft failure. The challenges indicate that the analytical tools of the classical dynamic programming methodology cannot be used in modeling our problem. However, AI techniques which have a broader scope of applicability such as machine learning (Sutton & Barto, 1998), online stochastic combinatorial optimization (Van Hentenryck & Bent, 2006), and hindsight optimization (Burns, Benton, Ruml, Yoon, & Do, 2012) can be investigated as potential approaches in future work.

**Extending the scheduling problem** Although our results are demonstrated for a specific class of scheduling problems where the only constraint is the repair capacity limit, our solution approach can be adapted for more complex scheduling problems. More specifically, the proposed MIP and CP algorithms for the static sub-problem can be easily extended to handle other types of scheduling constraints such as precedence constraints. However, modeling the problem via the decomposition approach would require additional effort. The existence of the precedence constraints among the repair activities of a failed aircraft makes the scheduling of different repair resources dependent. Therefore, a separate RSSP for each repair resource cannot be defined. One possible idea is to represent the scheduling problem as a single sub-problem where appropriate relaxation and Benders cut can be developed.

Taking another perspective, we decomposed the problem into stochastic long-term planning and deterministic short-term scheduling problems in DAMP and RSSP, respectively. In the long-term plan, we deal with the uncertainty using the known information on the probability of aircraft failures to trade-off the coverage between the flights. In the short-term schedule, we construct a feasible repair schedule to achieve the coverage as decided in the long-term plan. We then designed different rescheduling policies to investigate how the information revealed over time can be effectively used to adjust the repair schedule and change the long-term plan. We conclude that changing the long-term plan based on short-term information which cannot be completely incorporated in the long-term plan from the beginning significantly increases the utilization of the airplanes.

A typical hierarchical optimization approach does not deal with the interdependency between different levels of decision makings. However a communication technique as designed in this paper, capable of utilizing the short-term deterministic scheduling in the long-term stochastic planning is more likely to lead to higher system performance. In the problem addressed here, the assignment of the aircraft to the flights and the scheduling of the repair activities in the repair shop represent the long-term and the short-term reasoning, respectively. As a wide range of operational decisions can be viewed as integrated optimization problems, the pattern of the algorithms designed here might be applicable. Combining maintenance or inventory planning with job production scheduling is an example of the integrated operational problem where at the higher level of decision-making, the maintenance or inventory policy is determined whereas at the lower level, the jobs are scheduled (Terekhov, Doğru, Özen, & Beck, 2012; Aramon Bajestani, 2013). If the maintenance or inventory policy leads to an infeasible scheduling problem at the lower level, then the higher level needs to be informed with this information so that the inventory or maintenance decisions are adjusted. Therefore, the overall approach demonstrated here has applications to other problems typically solved with a hierarchical optimization approach.

## 6. Conclusion

In this paper, we address the problem of scheduling a dynamic repair shop in the context of aircraft fleet management. The goal is to maximize the flight coverage over a long-term by considering the repair capacity and the aircraft failures. The number of failed aircraft dynamically changes because of aircraft breakdowns. Our proposed solution solves the dynamic problem as successive static scheduling problems over shorter time periods. Several scheduling algorithms and different rescheduling policies are proposed to schedule the repair activities online with dynamic reaction to the aircraft failures. The length of the scheduling horizon and the frequency of rescheduling are the features defining our three policies.

The computational results show that an optimization approach using logic-based Benders decomposition, scheduling over a longer horizon, incorporating the mean of known information on aircraft failures, and adjusting the repair schedule as soon as new jobs enter the repair shop yield higher mean coverage and is a reasonable approach to balance against different uncertain scenarios.

Developing richer solution approaches to better handle the uncertainty is a challenging and interesting topic to pursue in future work. However, within our solution technique, establishing a formal framework to exactly determine how long into the future we should plan ahead and how quickly we should change our schedule based on the new information is also a very interesting direction for the future work. The answer of these two questions seems to be highly dependent on the arrival rate of the new information and the impact of the new information on the overall system performance. How to quantify these two values can be the starting point to provide solid responses to these two questions.

## Acknowledgments

This research was supported in part by the Natural Sciences and Engineering Research Council of Canada (NSERC) and the consortium members of Centre for Maintenance Optimization & Reliability Engineering (C-MORE). Thanks to Nima Safaei and Dragan Banjevic for introducing the problem to us and subsequent discussions. This paper is a combined and extended version of a conference paper (Aramon Bajestani & Beck, 2011b) and a workshop paper (Aramon Bajestani & Beck, 2011a) that have already appeared.